\documentclass[10pt,twocolumn,letterpaper]{article}

\usepackage[]{cvpr}      
\usepackage{algorithm}
\usepackage{algorithmic}

\usepackage{amsmath, amssymb, bm}
\usepackage{booktabs}        
\usepackage{multirow}        
\usepackage{adjustbox}       
\usepackage{makecell}        
\usepackage{float}           
\usepackage{pifont}
\usepackage{booktabs}

\usepackage{graphicx}

\usepackage[dvipsnames]{xcolor}
\definecolor{cvprblue}{rgb}{0.21,0.49,0.74}
\usepackage[pagebackref,breaklinks,colorlinks,allcolors=cvprblue]{hyperref}

\def\paperID{30361} 
\def\confName{CVPR}
\def\confYear{2026}

\newcommand{\mytracker}{ETCTrack}
\newcommand{\compressor}{Adaptive Token Compressor}
\newcommand{\encoder}{Hierarchical Interaction Encoder}
\newcommand{\cmark}{\ding{51}} 
\newcommand{\xmark}{\ding{55}} 

\usepackage[table]{xcolor}
\definecolor{mygreen}{RGB}{135,158,88}
\definecolor{myblue}{RGB}{61,66,183}

\title{An \underline{E}fficient \underline{T}oken \underline{C}ompression Framework for Visual Object Tracking}

\author{
    Weijing Wu\textsuperscript{\rm 1,2,3}, Qihua Liang\textsuperscript{\rm 1,2,3}\thanks{Corresponding Author}, Bineng Zhong\textsuperscript{\rm 1,2,3\raisebox{-1.5pt}{*}}, Haiying Xia\textsuperscript{\rm 1,2,3}, \\ Zhiyi Mo\textsuperscript{\rm 4}, Shuxiang Song\textsuperscript{\rm 1,2,3} \\
    \textsuperscript{\rm 1} Key Lab of Education Blockchain and Intelligent Technology, Ministry of Education,\\ Guangxi Normal University, Guilin, 541004, China \\
    \textsuperscript{\rm 2} Guangxi Key Lab of Multi-Source Information Mining and Security,\\
    Guangxi Normal University, Guilin, 541004, China \\
    \textsuperscript{\rm 3} University Engineering Research Center of Educational Intelligent Technology,\\ Guangxi Normal University, Guilin, 541004, China\\
    \textsuperscript{\rm 4} Guangxi Key Laboratory of Machine Vision and Intelligent Control,\\
    Wuzhou University, Wuzhou 543002, China.\\
    \tt\small wuweijing182@163.com, qhliang@gxnu.edu.cn,  bnzhong@gxnu.edu.cn, xhy22@mailbox.gxnu.edu.cn
}

\begin{document}
\maketitle

\begin{abstract}
Refining visual representations by eliminating their internal feature-level redundancy is crucial for simultaneously optimizing the performance and computational cost of models in visual tracking. To enhance their performance, many contemporary Transformer-based trackers leverage a larger number of historical template frames to capture richer spatio-temporal cues. However, this strategy leads to a massive number of input visual tokens. This creates two critical issues: it imposes a quadratic computational burden and can also degrade the tracker's overall performance. To bridge this gap, we propose a \textbf{compress-then-interact} tracking framework, \textbf{\mytracker}, that learns to efficiently compress template tokens from historical template frames into a robust target representation, moving beyond handcrafted rules. Our method first employs the \compressor{} to dynamically construct compact yet highly discriminative template tokens by filtering out redundant visual tokens. These refined template tokens are then processed by our Hierarchical Interaction Encoder to achieve a deep, adaptive interaction with the search features. Refined search features ensure subsequent precise target localization. Experiments on seven benchmarks demonstrate that our method outperforms current state-of-the-art trackers. ETCTrack-B224 reduces the number of template tokens by 60\%, leading to a 21.4\% reduction in MACs with only a 0.4\% drop in accuracy. The source code are available at https://github.com/PJD-WJ/ETCTrack.

\end{abstract}

\begin{figure}[t!]
   \centering
   \includegraphics[width=1.0\linewidth]{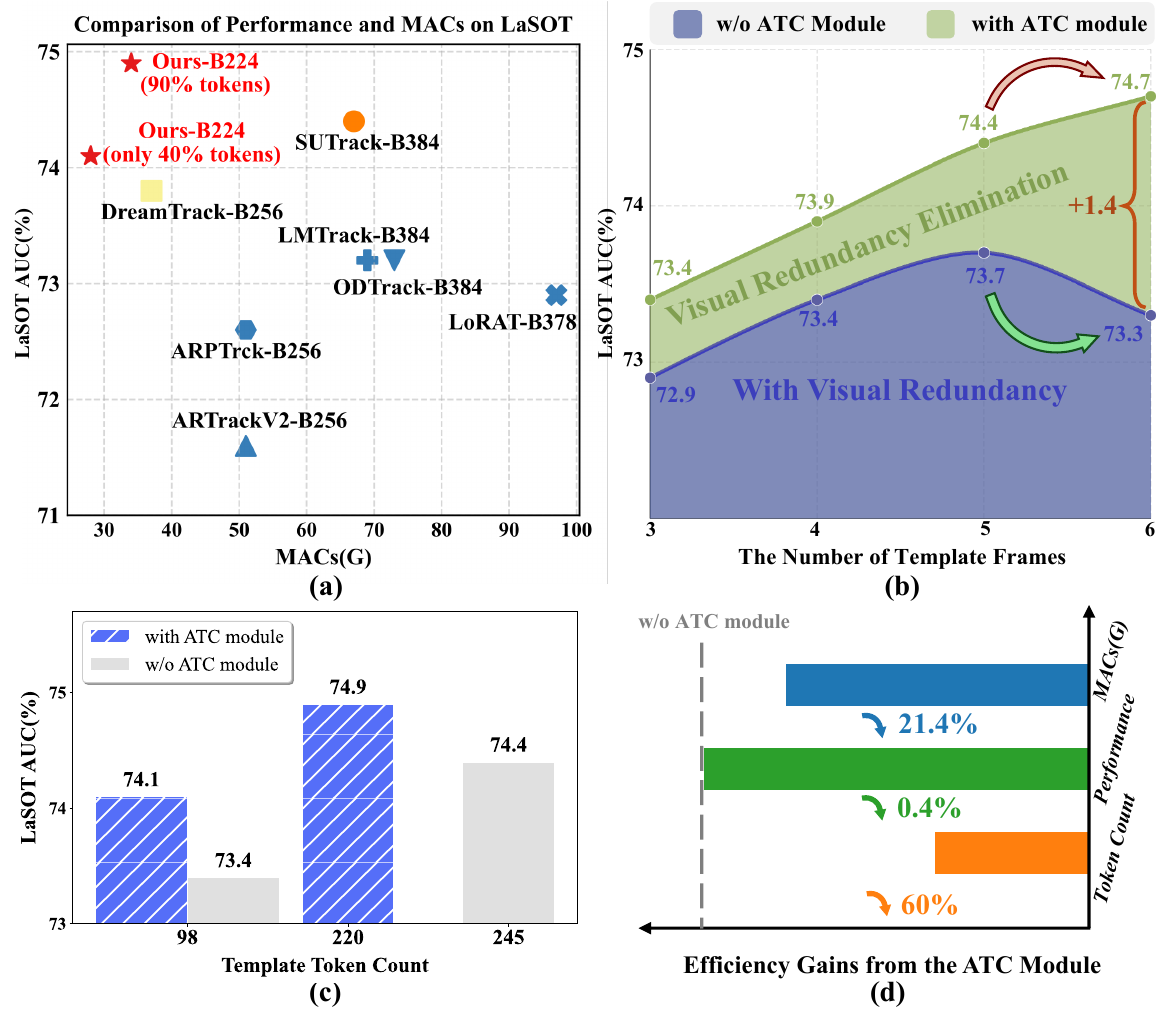}
   \caption{(a) Comparison of AUC and MACs of recent SOTA trackers. 
   (b) Baseline models decline after the 5th frame due to visual redundancy on LaSOT, but our ATC module overcomes this, enabling sustained performance gains with more frames.
   (c) Impact of ATC module and template token count on LaSOT AUC(\%).
   (d) Efficiency Gains from the ATC module.}
   \label{tracker_flops_and_auc_frame}
\end{figure}


\section{Introduction}
Visual tracking aims to locate a target throughout a video sequence using its initial state from the first frame as reference. Siamese network-based trackers \cite{SiamFC,SINT,SiamRPN} rose to prominence by reframing tracking as an efficient similarity-learning task between a template and a search region, achieving a strong balance of speed and accuracy. However, their similarity-matching design revealed limitations in handling complex appearance changes and contextual interactions, motivating the development of models with richer representations and stronger relational reasoning.

This pursuit led to the successful adaptation of Transformer architectures \cite{attention_is_all_you_need, hivit, vit} for visual object tracking, leveraging their powerful self-attention mechanisms to model global contextual information and complex feature interactions more effectively. Key methods such as those in \cite{TransT,Stark,OSTrack,SeqTrack,zheng2025towards} quickly established the prowess of this paradigm, setting new state-of-the-art performance benchmarks.
To further improve robustness against significant target appearance variations and for longe-term tracking, the method of multi-frame visual tracking emerged.
This approach, exemplified by methods \cite{Mixformer,ODTrack,TATrack,VideoTrack}, while aiming to build richer and more robust target representations from historical template frames, simultaneously introduces a critical new challenge: \textbf{a massive number of input visual tokens, which may introduce substantial visual redundancy that degrades model performance and incurs additional computational cost.}




Inspired by the core principle from Multimodal Large Language Models (MLLMs) \cite{Video-XL-Pro, InternVL-X, FocusLLaVA, Hybrid-Level-Instruction-Injection, nvidia2025_token_efficient} that mitigating visual redundancy is fundamental to balancing model capability with computational cost , we argue that this principle directly applies to a parallel challenge in multi-frame visual tracking. To empirically validate this assertion, we conducted a preliminary study, with results shown in Figure.\ref{tracker_flops_and_auc_frame}(b). For this study, we use the same number of template frames for both training and inference. We observe that when our representative baseline tracker, OSTrack \cite{OSTrack} equipped with a novel backbone \cite{itpn}, is augmented with an increasing number of historical template frames (\textcolor{myblue}{\textbf{blue line}}), its performance begins to decline significantly after the 5th frame. However, this performance ceiling is overcome when our ATC module (discussed further below) is introduced to eliminate the visual redundancy (\textcolor{mygreen}{\textbf{green line}}). \textbf{Our experiment confirms that visual redundancy is a critical bottleneck in multi-frame visual tracking.}

To alleviate the above problem, we propose a compress-then-interact framework, named \textbf{\mytracker{}}, which effectively compresses visual tokens from historical template frames to eliminate visual redundancy and enable more effective context modeling. The core of this framework consists of two components designed to work in synergy.
(1) The \compressor{} (ATC), a learnable module that acts as a crucial pre-processing stage before features enter our Hierarchical Interaction Encoder. It moves beyond handcrafted selection rules, which often rely on predefined metrics like attention maps or fixed spatial rules to measure token importance. Instead, our ATC module learns to dynamically evaluate the contextual importance of each token from historical template frames, guided directly by the final tracking objective. This allows it to construct a compact yet powerful subset of tokens that preserves the target's most discriminative features, providing a robust representation for interaction while reducing the computational cost (MACs).
(2) Our Hierarchical Interaction Encoder, constructed by stacking several Hierarchical Interaction Block (HIBlock) modules, then leverages compressed template representation to deeply interact with the search features. This is achieved through a multi-stage interaction process within each block, involving an context-aware enrichment of the template, followed by unified feature learning, and a final template-guided refinement of the search features. Finally, the interaction-enhanced search features are processed by the prediction head to estimate the target’s state.
As shown in Figure~\ref{tracker_flops_and_auc_frame}(a)(c)(d), our method achieves highest performance with the \textbf{lowest computational cost} among leading trackers, thanks to our core components.
To summarize, the main contributions of this work are as follows:

\begin{itemize}
    \item We propose a novel compress-then-interact framework for visual tracking that resolves the critical trade-off between performance and efficiency in multi-frame context modeling, establishing a new SOTA by outperforming leading trackers in both accuracy and computational cost.

    \item We introduce two key modules: the ATC module for efficiently compressing redundant tokens while preserving discriminative features, and the HIBlock for deep and effective contextual information interaction.
    
    \item Our method achieves SOTA results on seven challenging visual tracking benchmarks, including GOT10K, LaSOT, LaSOT$_{\text{ext}}$, TrackingNet, NfS, TNL2K, and OTB100.
\end{itemize}

\section{Related Work}

\subsection{Visual Tracking Architectures}
The advent of deep learning has revolutionized visual tracking, with the Siamese tracking framework \cite{SiamFC} emerging as a dominant architecture. Its compelling balance of accuracy and real-time efficiency spurred numerous works that aimed to significantly enhance overall tracking capabilities and robustness \cite{li2019siamrpn++, li2018high, fan2019siamese, fu2021stmtrack}. In recent years, the field has decisively shifted towards Transformer-based architectures \cite{attention_is_all_you_need,hivit,itpn}, valued for their superior capabilities in modeling global context and intricate feature interactions. This paradigm has evolved rapidly, from early hybrid models that integrated Transformer modules with CNNs \cite{TransT, Stark, gao2022robust, gao2023unambiguous}, to more recent unified, one-stream architectures that handle both feature extraction and interaction \cite{OSTrack, Mixformer, xue2025similarity, LMTrack,SiamPIN,MMTrack,sstrack}. Concurrently, the adoption of advanced hierarchical backbones like the Swin Transformer \cite{swintransformer, SwinTrack, song1} has further pushed performance boundaries, shaping the diverse landscape of modern state-of-the-art trackers \cite{SeqTrack, hu2025exploiting, li2025cadtrack, hu2025adaptive, hu2023transformer, hu2024toward, song2}.

\begin{figure*}[t]
   \centering
   \includegraphics[width=0.9\textwidth]{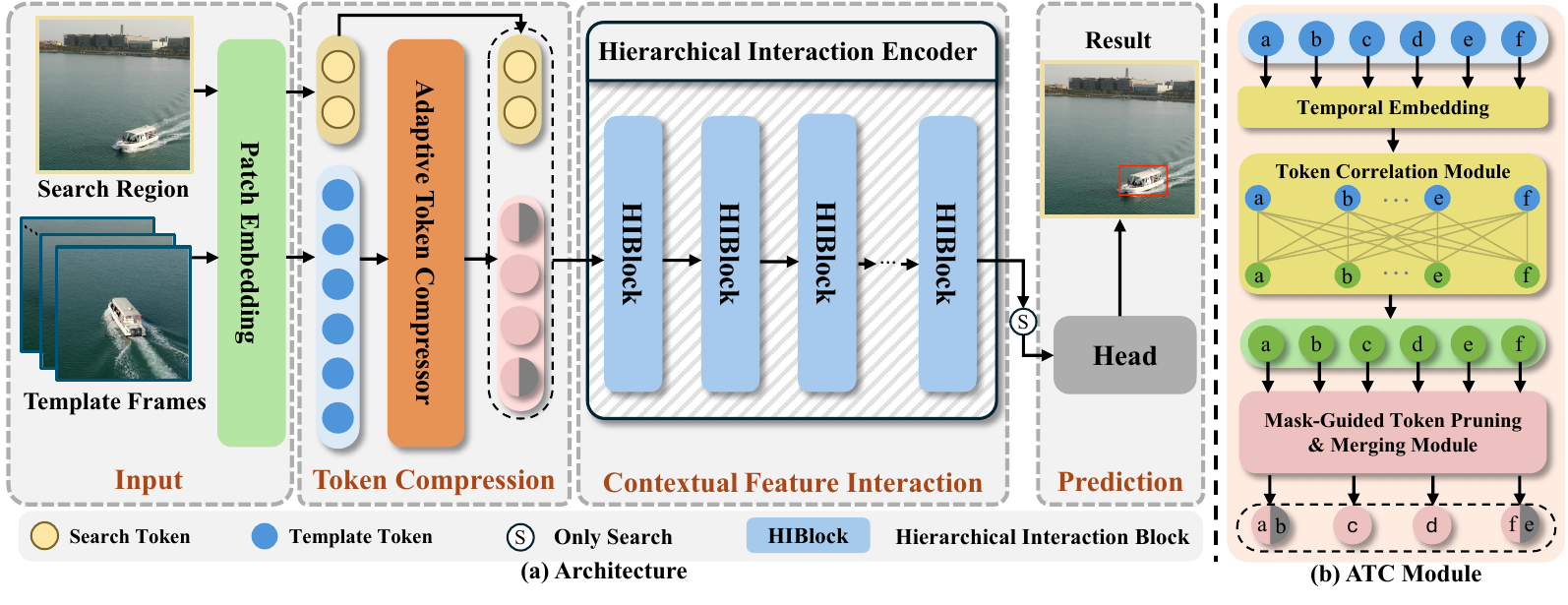}
   \caption{(a) \textbf{\mytracker{} Framework Architecture.} The process begins with our ATC module, which effectively compresses visual tokens from historical template frames to eliminate visual redundancy. These compressed tokens, along with search region tokens, are then fed into the Hierarchical Interaction Encoder for contextual feature interaction. Finally, the enhanced search features are sent to the Prediction Head to predict bounding box. (b) The detailed architecture of our core ATC module.
   }
   \label{architecture}
\end{figure*}


\subsection{Temporal Modeling in Visual Tracking}
Trackers have explored several strategies to better model temporal information. One such approach incorporates motion priors, using autoregressive models to predict locations from historical trajectories \cite{artrack, artrackv2}, but is prone to error accumulation during occlusions. Another strategy focuses on adaptive appearance modeling, where methods like \cite{evptrack, temtrack} learn to evolve the target's representation, replacing handcrafted update rules. The most prominent current approach, leveraged by advanced trackers \cite{TATrack, ODTrack, spmtrack}, is using historical template frames for a richer, more comprehensive spatio-temporal context. However, the increasing number of input visual tokens often imposes a quadratic computational burden and can also degrade tracking performance due to visual redundancy.

\begin{figure}[t]
   \centering
   \includegraphics[width=0.9\linewidth]{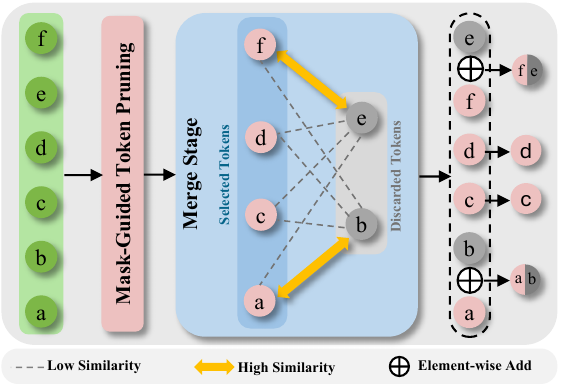}
   \caption{The Mask-Guided Token Pruning and Merging Module.}
   \label{similarity}
\end{figure}
\subsection{Visual Token Compression} 
The challenge of processing long video sequences with high visual redundancy has become a central research topic in Multimodal Large Language Models (MLLMs) \cite{blip2,flamingo,matryoshka,Moe-llava,liu2024improved}. The massive number of visual tokens from video inputs, coupled with the quadratic complexity of Transformers, has spurred the development of various visual token compression techniques, such as those based on handcrafted rules \cite{Madtp,Otterhd,Llava-prumerge,zhang2024tokenlevel}. While these rule-based solutions offer a straightforward approach, they are often sub-optimal as they rely on predefined criteria, such as importance scores derived from attention maps or fixed spatial rules, to evaluate token importance. Such non-learnable criteria may not align with the final tracking objective and can risk discarding crucial information. Therefore, this work focuses on developing an adaptive, learning-based framework to eliminate visual redundancy in tracking, aiming to simultaneously reduce computational cost and improve model performance.

\section{Our Method}
This section details the architecture of our proposed visual tracking framework, \mytracker{}. We first provide an overview of the overall pipeline, then elaborate on its three core components: the backbone, the ATC module, and the HIBlock. Finally, we describe the prediction head and the final training objective.


\subsection{Overview}
The framework of the \mytracker{} is demonstrated in Figure.\ref{architecture}(a).
The core components of our framework include an ATC module for eliminating visual redundancy, a HIBlock for contextual feature interaction, and a prediction head. At each tracking step, the framework processes historical template frames and the current search frame. These inputs are first partitioned into patches and embedded to generate template and search tokens. This collection of template tokens is then fed into our ATC module, which dynamically constructs a powerful subset by filtering out redundant or noisy information. This refined template tokens, along with the search tokens, is subsequently passed to Hierarchical Interaction Encoder for contextual feature interaction, where a multi-stage interaction process robustly correlates the target's essential features with search features. Finally, the resulting interaction-enhanced search tokens are fed into a prediction head to estimate the target's state.

\begin{figure}[t]
   \centering
   \includegraphics[width=0.9\linewidth]{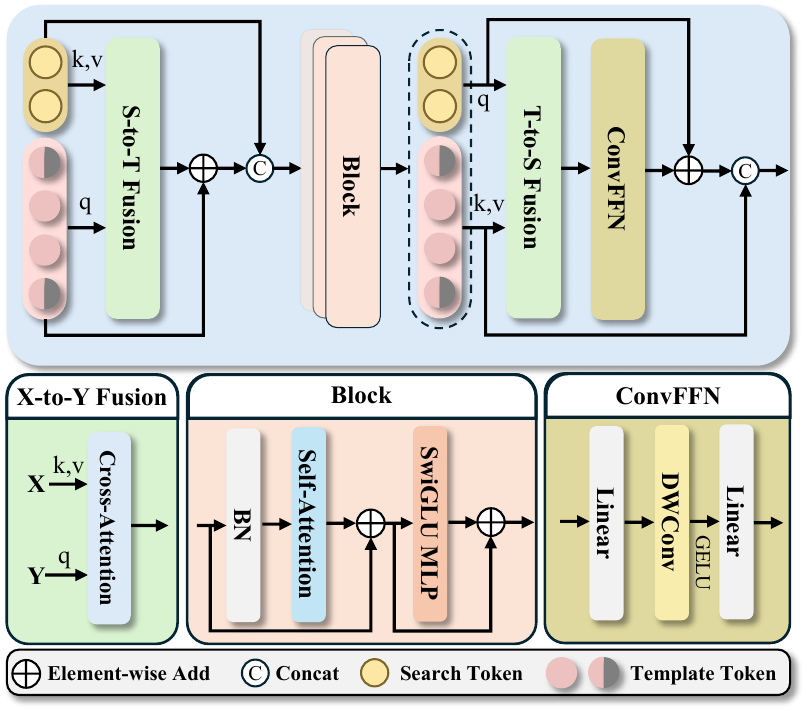}
   \caption{The structure of the Hierarchical Interaction Block.}
   \label{HIBlock}
\end{figure}
\subsection{\compressor}
The ATC module is a core component of our framework, designed to eliminate visual redundancy in multi-frame tracking. Inspired by similar challenges in Multimodal Large Language Models (MLLMs) \cite{blip2,FocusLLaVA,flamingo}, the ATC module moves beyond handcrafted rules by learning to intelligently construct a compact yet highly representative subset of tokens from historical template frames. It employs a powerful global attention mechanism to evaluate the contextual importance of each token, allowing it to filter out noisy and redundant information. This process yields a highly discriminative template representation that simultaneously enhances tracking robustness while reducing computational cost.

As shown in Figure~\ref{architecture}(b), the module takes template features $\bm{F}_{\text{z}} \in \mathbb{R}^{(T \cdot L) \times C}$ as input. First, we restore the explicit spatio-temporal structure by reshaping the input into $\bm{Z}_{\text{p}} \in \mathbb{R}^{T \times L \times C}$, where $T$ is the number of historical template frames, $L$ is the number of tokens per frame, and $C$ is the feature dimension. To distinguish tokens from different frames, a learnable temporal positional embedding $\bm{E}_{\text{temp}} \in \mathbb{R}^{T \times 1 \times C}$ is then added to $\bm{Z}_{\text{p}}$. The resulting vector is flattened along its temporal and spatial dimensions to form $\bm{Z}_{\text{temp}} \in \mathbb{R}^{(T \cdot L) \times C}$.
Next, $\bm{Z}_{\text{temp}}$ is processed by our proposed Token Correlation Module (TCM), which consists of $N_{atc}$ stacked self-attention layers. The TCM, through its global self-attention mechanism, deeply interacts with all tokens from template frames, which already have temporal positional information embedded. This allows the TCM to model the comprehensive spatio-temporal relationships across the template frames, enabling it to understand complex correlations between all template tokens and identify the most discriminative and informative parts. This process yields the contextualized feature representations $\bm{Z}_{\text{context}}$:
\begin{equation}
    \bm{Z}_{\text{context}} = \text{TCM}(\bm{Z}_{\text{temp}}) \in \mathbb{R}^{(T \cdot L) \times C}
    \label{eq:tcm_output}
\end{equation}
After obtaining $\bm{Z}_{\text{context}}$, we introduce the Mask-Guided Token Pruning module to evaluate token significance, as illustrated in Figure~\ref{similarity}. We employ a fixed random projection to generate token importance scores $\bm{S}$, acting as a randomized token selection mechanism to enhance the model's capability in learning efficient similarity-based merging. Based on the descending order of $\bm{S}$, the tokens are partitioned into a preserved target set $\bm{\mathcal{A}} = \{\bm{a}_j\}_{j=1}^{K_{\text{target}}}$ (high score) and a redundant source set $\bm{\mathcal{B}} = \{\bm{b}_i\}_{i=1}^{K_{\text{merge}}}$ (low score). The number of merged tokens is determined by a predefined keep rate $r \in (0, 1)$, yielding $K_{\text{target}} = \lfloor r \cdot (T \cdot L) \rfloor$ and $K_{\text{merge}} = (T \cdot L) - K_{\text{target}}$.

To prevent semantic information loss from directly discarding $\bm{\mathcal{B}}$, we propose a Guided Similarity Merging strategy. Each source token $\bm{b}_i \in \bm{\mathcal{B}}$ is absorbed into its most semantically similar target token $\bm{a}_j \in \bm{\mathcal{A}}$ via greedy cosine similarity matching:
\begin{equation}
    \bm{a}'_j = \bm{a}_j + \sum_{i \in \Omega_j} \bm{b}_i
    \label{eq:gsm_fusion}
\end{equation}
where $\Omega_j = \big\{i \mid j = \arg\max_{k} (\bm{b}_i \cdot \bm{a}_k) / (\|\bm{b}_i\|_2 \|\bm{a}_k\|_2) \big\}$ denotes the indices of source tokens assigned to $\bm{a}_j$. The updated feature set $\bm{F}_{\text{comp}} = \{\bm{a}'_j\}_{j=1}^{K_{\text{target}}} \in \mathbb{R}^{K_{\text{target}} \times C}$ effectively eliminates redundancy while yielding a highly efficient, context-aware template representation.





\subsection{\encoder}
\noindent\textbf{Backbone.}
The backbone architecture for state-of-the-art visual trackers \cite{ODTrack, LMTrack}, has increasingly shifted towards ViT \cite{vit} for their powerful feature extraction capabilities.
First, fine-grained recognition tasks like visual tracking require hierarchical features to capture information at various scales. However, the majority of standard pre-trained models are based on plain ViT architectures that produce a single-scale, non-hierarchical feature map, which is sub-optimal for robustly localizing targets of different sizes.
To address this, Fast-iTPN \cite{itpn} was chosen as our backbone.

We take historical template frames $\bm{Z} \in \mathbb{R}^{T \times 3 \times H_z \times W_z}$ and a search region $\bm{X} \in \mathbb{R}^{3 \times H_x \times W_x}$ as inputs. Both $\bm{Z}$ and $\bm{X}$ are first downsampled using convolutional layers with a stride of 4. The downsampled features are then passed through MLP layers and two convolutional merging layers, which segment the inputs into non-overlapping patches.
This process yields patch embeddings for the historical template frames and the search region, denoted as $\bm{F}_z \in \mathbb{R}^{N_z \times C}$ and $\bm{F}_x \in \mathbb{R}^{N_x \times C}$, respectively, where $N_z = \frac{T \cdot H_z \cdot W_z}{16^2}, \quad N_x = \frac{H_x \cdot W_x}{16^2}.$ The resulting patches are then concatenated along the spatial dimension to form $\bm{F}_{zx} \in \mathbb{R}^{L \times C}$, where $L = N_z + N_x$. 
$\bm{F}_{zx}$ subsequently serves as the input to our core Hierarchical Interaction Encoder, where template and search tokens undergo a deep, multi-stage interaction process.

\noindent\textbf{Hierarchical Interaction Block.}
The principle of jointly performing feature extraction and relation modeling within a backbone, as demonstrated by effective tracker\cite{OSTrack}, has proven to be a powerful approach for visual tracking. By processing template and search region tokens together, these methods enable rich feature interaction. However, the extent and nature of this interaction can be further deepened and made more explicit. While joint self-attention allows for implicit correlation, a more structured, multi-stage interaction mechanism is beneficial to fully allow the template representation to guide the search process and, in turn, for the search context to refine the template representation. This motivates the design of our HIBlock.

The detailed structure of the HIBlock is shown in Figure~\ref{HIBlock}. The HIBlock is the central core component of \mytracker{}. It takes the compressed template tokens from the ATC module and the search tokens as input. The HIBlock is designed to perform a deep, asymmetric interaction process. This hierarchical and multi-stage design ensures a highly adaptive exchange of information between the template tokens and the search tokens.

\begin{table*}[t!]
    \centering
    \begin{adjustbox}{valign=c,max width=\textwidth}
        \fontsize{10}{12}\selectfont
        \begin{tabular}{r|c|ccc|ccc|ccc|ccc|cc}
        \toprule
        \multicolumn{1}{c|}{\multirow{2}{*}{Method} }
        & \multicolumn{1}{c|}{\multirow{2}{*}{Source}} 
        & \multicolumn{3}{c|}{GOT-10k$^*$} 
        & \multicolumn{3}{c|}{LaSOT} 
        & \multicolumn{3}{c|}{LaSOT$_{ext}$} 
        & \multicolumn{3}{c|}{TrackingNet}
        & \multicolumn{2}{c}{TNL2k} \\
        \cline{3-16}
        && AO & SR$_{0.5}$ & SR$_{0.75}$ & AUC & P$_{norm}$ & P & AUC & P$_{norm}$ & P & AUC & P$_{norm}$ & P & AUC & P \\
        \midrule

        \rowcolor{orange!15}
        \textbf{{\mytracker}-B224} & Ours & \textbf{79.2} & \textbf{90.2} & \underline{78.6} & \textbf{74.9} & \textbf{85.1} & \textbf{82.7} & \textbf{54.6} & \textbf{65.9} & \textbf{62.4} & \underline{86.0} & \underline{90.6} & \underline{85.6} & \textbf{61.3} & \textbf{65.9} \\

        \rowcolor{blue!10}
        \textbf{{\mytracker}-S224} & Ours & 75.8 & 86.6 & 75.0 & 73.4 & 83.5 & 81.1 & \underline{53.7} & \underline{65.0} & \underline{61.1} & 85.4 & 90.1 & 84.8 & 60.0 & 63.4 \\

        \rowcolor{gray!30}
        \textbf{{\mytracker}-T224} & Ours & 74.9 & 85.6 & 72.6 & 71.8 & 82.1 & 78.9 & 52.5 & 63.9 & 59.4 & 84.6 & 89.3 & 83.2 & 58.3 & 61.0 \\

        \midrule

        DreamTrack-B256 \cite{DreamTrack} & CVPR25 & 77.5 & 87.1 & 74.2 & 73.8 & 83.4 & 80.6 & 53.1 & 64.1 & 59.8 & 85.8 & 90.0 & 85.3 & \underline{60.4} & 63.2 \\
        ARPTrack-B256 \cite{ARPTrack} & CVPR25 & 77.7 & 87.3 & 74.3 & 72.6 & 81.4 & 78.5 & 52.0 & 62.9 & 58.7 & 85.5 & 90.0 & 85.3 & - & - \\
        MCITrack-B224\dag \cite{MCITrack} & AAAI25 & \underline{77.9} & \underline{87.5} & \textbf{78.8} & \underline{74.0} & \underline{83.6} & \underline{81.4} & 52.0 & 62.5 & 59.0 & \textbf{86.4} & \textbf{91.0} & \textbf{86.3} & 60.3 & \underline{64.4} \\
        TemTrack-B256 \cite{temtrack} & AAAI25 & 74.9 & 84.8 & 71.7 & 72.0 & 82.1 & 79.1 & 52.4 & 63.3 & 60.2 & 84.3 & 88.8 & 83.5 & 58.8 & - \\
        ARTrackV2-B256 \cite{artrackv2} & CVPR24 & 75.9 & 85.4 & 72.7 & 71.6 & 80.2 & 77.2 & 50.8 & 61.9 & 57.7 & 84.9 & 89.3 & 84.5 & 59.2 & - \\
        VideoTrack\cite{VideoTrack1} & CVPR23 & 72.9 & 81.9 & 69.8 & 70.2 & - & 76.4 & - & - & - & 83.8 & 88.7 & 83.1 & - & - \\
        MixFormer-22k\cite{Mixformer} & CVPR22 & 70.7 & 80.0 & 67.8 & 69.2 & 78.7 & 74.7 & - & - & - & 83.1 & 88.1 & 81.6 & - & - \\
        OSTrack-256\cite{OSTrack} & ECCV22 & 71.0 & 80.4 & 68.2 & 69.1 & 78.7 & 75.2 & 47.4 & 57.3 & 53.3 & 83.1 & 87.8 & 82.0 & 54.3 & - \\
        TransT \cite{TransT} & CVPR21 & 67.1 & 76.8 & 60.9 & 64.9 & 73.8 & 69.0 & - & - & - & 81.4 & 86.7 & 80.3 & 50.7 & 51.7 \\
        SiamRPN++\cite{li2019siamrpn++} & CVPR19 & 51.7 & 61.6 & 32.5 & 49.6 & 56.9 & 49.1 & 34.0 & 41.6 & 39.6 & 73.3 & 80.0 & 69.4 & 41.3 & 41.2 \\
        SiamFC \cite{SiamFC} & ECCVW16 & 34.8 & 35.3 & 9.8 & 33.6 & 42.0 & 33.9 & 23.0 & 31.1 & 26.9 & - & - & - & 29.5 & 28.6 \\

        \midrule

        \multicolumn{16}{l}{\textit{Some Trackers with Higher Resolution}} \\

        \midrule

        OSTrack-384\cite{OSTrack} & ECCV22 & 73.7 & 83.2 & 70.8 & 71.1 & 81.1 & 77.6 & 50.5 & 61.3 & 57.6 & 83.9 & 88.5 & 83.2 & 55.9 & - \\
        F-BDMTrack-384\cite{F-BDMTrack}&ICCV23  & 75.4 & 84.3 & 72.9 & 72.0 & 81.5 & 77.7 & 50.8 & 61.3 & 57.8 & 84.5 & 89.0 & 84.0 & 57.8 & 59.4 \\
        LoRAT-B378\cite{hiptrack} & ECCV24 & 73.7 & 82.6 & 72.9 & 72.9 & 81.9 & 79.1 & 53.1 & 64.8 & 60.6 & 84.2 & 88.4 & 83.0 & 59.9 & 63.7 \\
        HIPTrack-B384\cite{hiptrack} & CVPR24 & 77.4 & \underline{88.0} & 74.5 & 72.7 & 82.9 & 79.5 & 53.0 & 64.3 & 60.6 & 84.5 & 89.1 & 83.8 & - & - \\
        ARTrackV2-B384 \cite{artrackv2} & CVPR24 & 77.5 & 86.0 & 75.5 & 73.0 & 82.0 & 79.6 & 52.9 & 63.4 & 59.1 & 85.7 & 89.8 & 85.5 & - & - \\
        ODTrack-B384 \cite{ODTrack} & AAAI24 & 77.0 & 87.9 & 75.1 & 73.2 & 83.2 & 80.6 & 52.4 & 63.9 & 60.1 & 85.1 & 90.1 & 84.9 & 60.9 & - \\
        TemTrack-B384 \cite{temtrack} & AAAI25 & 76.1 & 84.9 & 74.4 & 73.1 & 83.0 & 80.7 & 53.4 & 64.8 & 61.0 & 85.0 & 89.3 & 84.8 & - & - \\
        SPMTrack-B384 \cite{spmtrack} & CVPR25 & 76.5 & 85.9 & 76.3 & 74.9 & 84.0 & \underline{81.7} & - & - & - & 86.1 & 90.2 & 85.6 & \underline{62.0} & 66.7 \\
        DreamTrack-B384 \cite{DreamTrack} & CVPR25 & \underline{78.3} & 87.9 & \underline{76.6} & \underline{75.0} & \underline{84.2} & \underline{81.7} & \underline{54.5} & \underline{65.3} & \underline{61.1} & \underline{86.5} & \underline{90.6} & \underline{85.9} & 61.2 & \underline{67.4} \\

        \midrule

        \rowcolor{orange!15}
        \textbf{{\mytracker}-B384} & Ours & \textbf{80.1} & \textbf{89.5} & \textbf{80.7} & \textbf{75.9} & \textbf{85.9} & \textbf{84.0} & \textbf{55.1} & \textbf{66.2} & \textbf{62.5} & \textbf{87.3} & \textbf{91.7} & \textbf{88.2} & \textbf{63.0} & \textbf{68.4} \\

        \bottomrule    
        \end{tabular}
    \end{adjustbox}
    \caption{
        Performance comparison with SOTA trackers on the test sets of GOT-10k \cite{got10k}, LaSOT \cite{lasot}, $\rm LaSOT_{ext}$ \cite{lasot_ext}, TrackingNet \cite{trackingnet}, and TNL2K \cite{tnl2k}. For GOT-10k, an asterisk (*) denotes models trained exclusively on its training set. MCITrack-B224\dag~was locally trained under identical model, GPU, and configuration conditions. The top two results are highlighted with \textbf{bold} and \underline{underlined} fonts, respectively.
    }
    \label{performance_architecture}
\end{table*}
The detailed workflow of the HIBlock is as follows. For a single sample, the block receives the compressed template tokens $\bm{F}_{\text{comp}} \in \mathbb{R}^{K \times C}$ and the search tokens $\bm{F}_{\text{x}} \in \mathbb{R}^{N_x \times C}$. First, to make the template representation aware of the current search context, an cross-attention fusion is performed where the template tokens $\bm{F}_{\text{comp}}$ act as $Q$, while the search tokens $\bm{F}_{\text{x}}$ serve as $K$ and $V$. This step yields the context-aware template tokens, denoted as $\bm{F}'_{\text{comp}}$. These are then concatenated with the original search tokens and fed into a stack of $M$ standard backbone blocks \cite{itpn}, for deep, joint feature modeling. The output of this stage, $\bm{F}_{\text{block}}$ is subsequently split back into its template and search tokens, yielding deeply encoded features $\bm{F}''_{\text{comp}}$ and $\bm{F}'_{\text{x}}$, respectively. Next, to explicitly guide the search process, a second cross-attention is performed where the search tokens $\bm{F}'_{\text{x}}$ query the deeply encoded template tokens $\bm{F}''_{\text{comp}}$, resulting in the enhanced search tokens $\bm{F}''_{\text{x}}$. Finally, these enhanced features are passed through a Convolutional Feed-Forward Network (ConvFFN) for final refinement, producing the output features $\bm{F}_{\text{out}} \in \mathbb{R}^{N_x \times C}$ which are then sent to the prediction head. The process is summarized as follows:
\begin{equation}
   \label{eq:combined_equations} 
   \begin{aligned}
       \bm{F}'_{\text{comp}} &= \bm{F}_{\text{comp}} + \text{CrossAttention}(\bm{F}_{\text{comp}}, \bm{F}_{\text{x}}, \bm{F}_{\text{x}}), \\
       \bm{F}_{\text{block}} &= \text{BackboneBlocks}(\text{Concat}(\bm{F}'_{\text{comp}}, \bm{F}_{\text{x}})), \\
       \bm{F}''_{\text{x}} &= \bm{F}'_{\text{x}} + \text{CrossAttention}(\bm{F}'_{\text{x}}, \bm{F}''_{\text{comp}}, \bm{F}''_{\text{comp}}), \\
       \bm{F}_{\text{out}} &= \text{ConvFFN}(\bm{F}''_{\text{x}}).
   \end{aligned}
\end{equation}

\begin{table*}[t]
   \setlength{\aboverulesep}{0pt}
   \setlength{\belowrulesep}{0pt}
   \renewcommand{\arraystretch}{1.1}
   \centering
   \begin{adjustbox}{valign=c,max width=\textwidth}
   \setlength\tabcolsep{4pt} 
   \fontsize{10}{10.8}\selectfont

   \begin{tabular}{l|c c c c c c c|>{\columncolor{orange!15}}c >{\columncolor{orange!15}}c}
   \toprule
   \textbf{} & DreamTrack-B384 & SPMTrack-B384 & ARTrackV2-L384 & HIPTrack-B384 & ODTrack-B384 & LoRAT-L378 & SeqTrack-B384 
   & \textbf{\mytracker-B224} 
   & \textbf{\mytracker-B384} \\
   \midrule
   \textbf{NfS} & 66.1 & 67.4 & 68.4 & 68.1 & - & 66.7 & 66.7 & \textbf{71.3} & \underline{71.2} \\
   \textbf{OTB100} & 72.0 & 72.7 & - & 71.0 & 72.3 & -    & -    & \underline{73.3} & \textbf{73.6} \\
   \bottomrule
   \end{tabular}

   \end{adjustbox}
   \caption{Comparison with state-of-the-art methods on NfS \cite{nfs} and OTB100 \cite{otb100} benchmarks in AUC score. The top two results are highlighted with \textbf{bold} and \underline{underlined} fonts, respectively.}
   \label{tab:nfs_otb100}
\end{table*}


\begin{table}[h!]
   \centering
   \setlength\tabcolsep{3.5pt} 
   \renewcommand{\arraystretch}{0.85}
   \fontsize{8}{10.8}\selectfont

   \begin{tabular}{c|cc|cc|cc|c}
   \toprule
   \multicolumn{1}{c|}{\multirow{2}{*}{\#}} & 
   \multicolumn{1}{c}{\multirow{2}{*}{ATC Module}} & 
   \multicolumn{1}{c|}{\multirow{2}{*}{HIBlock}} & 
   \multicolumn{2}{c|}{LaSOT} & 
   \multicolumn{2}{c|}{TNL2K} & 
   \multicolumn{1}{c}{FLOPs} \\
   \cline{4-7}
   & & & AUC & P$_{norm}$ & AUC & P$_{norm}$ & (G)\\
   \midrule

   \rowcolor{gray!8}
   1 & \xmark & \xmark & 73.7 & 83.7 & 59.9 & 77.7 & 33 \\

   \rowcolor{gray!8}
   2 & \cmark & \xmark & 74.4 & 84.6 & 60.6 & 78.3 & 32 \\
   
   \rowcolor{gray!8}
   3 & \xmark & \cmark & 74.4 & 84.5 & 60.5 & 78.0 & 36 \\

   \rowcolor{orange!15}
   4 & \cmark & \cmark & \textbf{74.9} & \textbf{85.1} & \textbf{61.3} & \textbf{78.8} & 34 \\

   \bottomrule
   \end{tabular}
   \caption{Ablation studies on ATC module and HIBlock.}
   \label{ablation_compression_interation}
\end{table}



\begin{figure}[t!]
   \centering
   \includegraphics[width=0.9\linewidth]{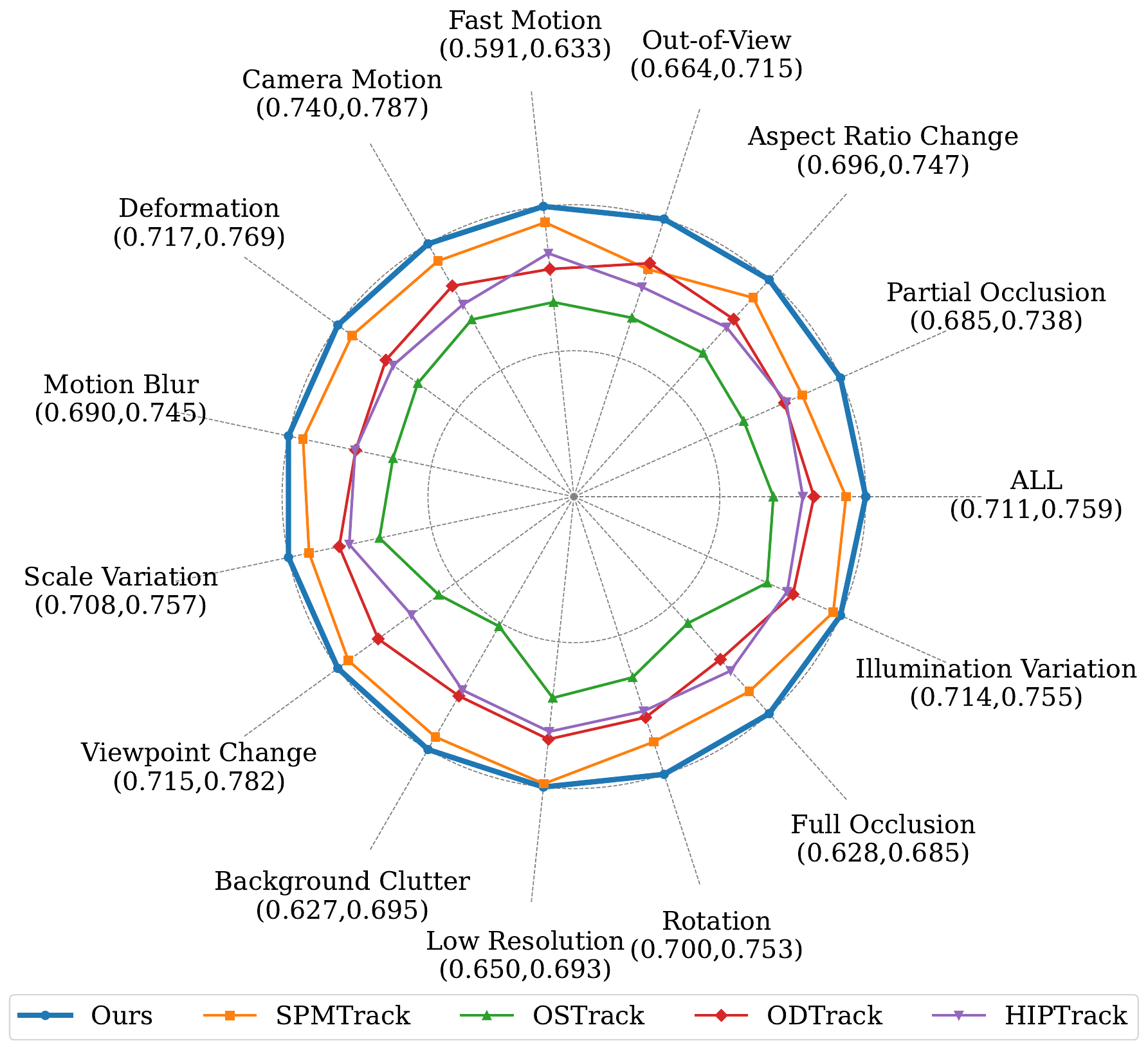}
   \caption{AUC scores of different attributes on LaSOT \cite{lasot}.}
   \label{eao}
\end{figure}

\subsection{Head and Loss}
The output features $\bm{F}_{\text{out}}$ from the Hierarchical Interaction Encoder are forwarded to a Fully Convolutional Network (FCN), which comprises $L$ stacked Conv-BN-ReLU layers in each output branch. The FCN generates three types of outputs: a target classification score map of size $\mathbb{R}^{\frac{H_x}{P} \times \frac{W_x}{P}}$, an offset map of size $\mathbb{R}^{2 \times \frac{H_x}{P} \times \frac{W_x}{P}}$ to compensate for discretization errors introduced by downsampling, and normalized bounding box dimensions of size $\mathbb{R}^{2 \times \frac{H_x}{P} \times \frac{W_x}{P}}$.


During the training phase, the classification and regression objectives are jointly optimized. For the classification task, which aims to distinguish the target from the background, we adopt the weighted focal loss \cite{focal_loss}. For bounding box regression, which is crucial for precise localization, we employ a combination of the $L_1$ loss and the generalized IoU loss \cite{giou} on the predicted box coordinates. The comprehensive optimization objective is formulated as:
\begin{equation}
    L = L_{\text{cls}} + \lambda_{\text{iou}} L_{\text{iou}} + \lambda_{L_1} L_1,
\end{equation}
where the weights $\lambda_{L_1}$ and $\lambda_{\text{iou}}$ are set to 5 and 2.





\section{Experiment}

\subsection{Implementation Details} 
\noindent\textbf{Training.}
We use Fast-iTPN \cite{itpn} model as our backbone. The training data includes LaSOT \cite{lasot}, GOT-10k \cite{got10k}, TrackingNet \cite{trackingnet}, COCO \cite{coco} and VastTrack \cite{vastrack}. We employ the AdamW to optimize the network parameters with initial learning rate of $2 \times 10^{-5}$ for the backbone, $2 \times 10^{-4}$ for the rest, and set the weight decay to $10^{-4}$. We set the training epochs to 300 epochs. 60{,}000 search images are randomly sampled in each epoch. The learning rate drops by a factor of 10 after 240 epochs. The model is conducted on a server with four 80GB Tesla A800 GPUs, using a batch size of 16, where each batch consists of 2 search images and 5 template images.

\begin{table}[t!]
\centering
\setlength{\tabcolsep}{0.1pt} 
\scalebox{0.8}{
\begin{tabular}{c|ccccc}
\toprule
Keep Ratio($r$) & LaSOT$\uparrow$ & LaSOT$_{\text{ext}}$$\uparrow$ & NfS$\uparrow$ & TNL2K$\uparrow$ & MACs(G)\textcolor{olive}{$\downarrow$} \\
\midrule
\rowcolor{gray!8}
$1.0$ (w/o ATC) & 74.4 & 53.5 & 71.2 & 60.5 & 35.9 \\
\rowcolor{orange!15}
$0.9$ & \textbf{74.9}(\textbf{\textcolor{red}{+0.5}}) & \textbf{54.6}(\textbf{\textcolor{red}{+1.1}}) & \textbf{71.3}(\textbf{\textcolor{red}{+0.1}}) & \textbf{61.3}(\textbf{\textcolor{red}{+0.8}}) & 34.1 \\
\rowcolor{gray!8}
$0.7$ & 74.5(\textbf{\textcolor{red}{+0.1}}) & 53.8(\textbf{\textcolor{red}{+0.3}}) & 71.1(\textbf{\textcolor{OliveGreen}{-0.1}}) & 60.8(\textbf{\textcolor{red}{+0.3}}) & 31.7 \\
\rowcolor{gray!8}
$0.5$ & 74.3(\textbf{\textcolor{OliveGreen}{-0.1}}) & 54.0(\textbf{\textcolor{red}{+0.5}}) & 71.3(\textbf{\textcolor{red}{+0.1}}) & 60.6(\textbf{\textcolor{red}{+0.1}}) & \textbf{28.2} \\
\bottomrule
\end{tabular}}
\caption{Performance vs. keep ratio ($r$).}
\end{table}
\noindent\textbf{Inference.}
During inference, our core strategy is to compress template information using the ATC module, which retains only the most representative tokens based on a token keep ratio($r$) of $0.9$ to create a potent yet efficient representation. Consistent with our training process, this mechanism is applied to 5 historical template frames that are processed alongside the current search region. Furthermore, to effectively mitigate error accumulation in long-term tracking, we employ a dynamic memory bank that is selectively updated. A new frame is added only if its maximum classification score surpasses a confidence threshold $\tau$, ensuring the bank stores only reliable, high-quality historical template frames for robust long-term tracking.

\begin{table}[t]
   \centering
   \setlength\tabcolsep{5.5pt} 
   \renewcommand{\arraystretch}{0.9} 
   \fontsize{8}{10.8}\selectfont
   \begin{tabular}{c|c c}
   \toprule
   TCM Architecture & AUC & Speed(fps) \\
   \midrule
   \rowcolor{gray!8}
   ViT-B(MAE pre-trained, 12 layers) & 74.7 & 44 \\
   \rowcolor{gray!8}
   ViT-B(from scratch, 12 layers) & 74.2 & 44 \\
   \rowcolor{gray!8}
   Transformer Encoder(8 layers) & 74.9 & 46 
   \\
   \rowcolor{orange!15}
   \textbf{Fast-ITPN(The first 8 layers)} & \textbf{74.9} & \textbf{48} \\
   \bottomrule
   \end{tabular}
   \caption{Ablation on Different TCM Architecture on LaSOT.}
   \label{TCM_ScoreProjector_Architecture}
\end{table}

\begin{figure}[t!]
   \centering
   \includegraphics[width=0.75\linewidth]{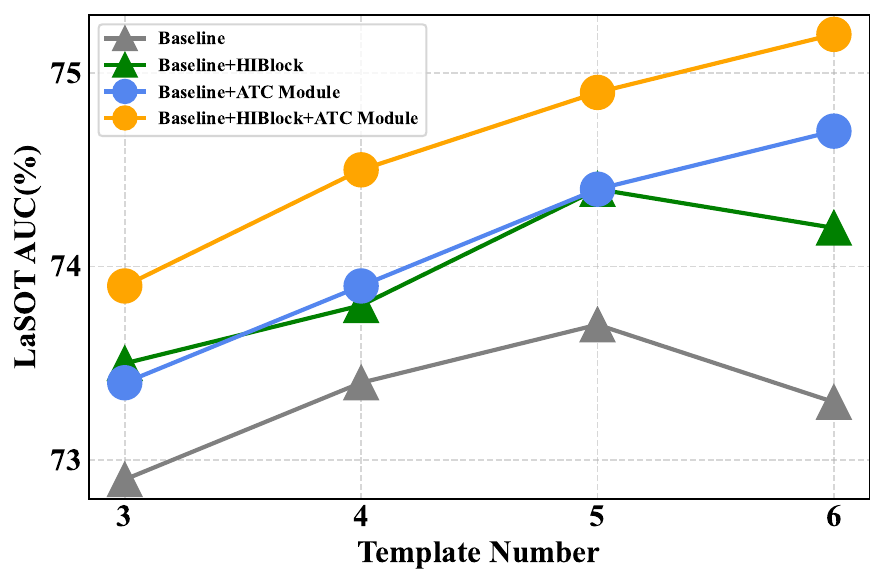}
   \caption{LaSOT AUC vs. template frames for our variants.}
   \label{ratio_frame}
\end{figure}




\subsection{Comparison with State-of-the-Art Trackers}
To demonstrate the effectiveness of \mytracker, we compare it against state-of-the-art trackers on seven diverse benchmarks, including GOT-10k, TrackingNet, LaSOT, LaSOT$_{ext}$, TNL2K, NfS, and OTB100.

\noindent\textbf{GOT-10k \cite{got10k}.} 
GOT-10K. The GOT-10k benchmark is a large-scale tracking dataset comprising over 10,000 video sequences. Its strict protocol mandates that trackers use only its designated training set for training, a guideline we adhered to for our framework. As shown in Table~\ref{performance_architecture}, our \mytracker{}-B224 achieved an AO score of 79.2\%, surpassing the second-place MCITrack-B224 (77.9\% AO). Our proposed method significantly outperforms previous SOTA trackers, even achieving higher performance than all higher-resolution trackers.

\noindent\textbf{LaSOT \cite{lasot}.} 
LaSOT is a large-scale, long-term tracking benchmark that includes 1,120 sequences for training and 280 for testing, known for its challenging scenarios. As presented in Table~\ref{performance_architecture} and Figure~\ref{eao}, our \mytracker{}-B384 achieves a new state-of-the-art result on this demanding benchmark. Specifically, our method surpasses the strong competitor DreamTrack-B384 by 0.9\%, 1.7\%, and 2.1\% in terms of AUC, $P_{\text{norm}}$, and P scores, respectively.



\noindent\textbf{LaSOT$_{ext}$ \cite{lasot_ext}.} This benchmark is an extension of LaSOT, introducing an additional 150 long-term sequences with numerous challenges, such as fast-moving small objects. As reported in Table~\ref{performance_architecture}, our method outperforms other trackers by a substantial margin, achieving the highest scores across all metrics (AUC, $P_{\text{norm}}$, and P). For instance, our method achieves an AUC of 55.1\%, surpassing the previous best, DreamTracker-B384, by 0.6\%.

\noindent\textbf{TrackingNet \cite{trackingnet}.} TrackingNet is a large-scale tracking dataset with over 30,000 training sequences and 511 test sequences. This benchmark focuses on challenges encountered when tracking objects in the wild, such as background clutter, full occlusion, and low resolution. The results on the TrackingNet test set are shown in Table~\ref{performance_architecture}. Our \mytracker{} achieves an AUC score of 87.3\%, outperforming the strong runner-up DreamTrack-B384 (86.5\% AUC) and demonstrating the robustness of our method in unconstrained environments.

\noindent\textbf{TNL2K \cite{tnl2k}.} 
TNL2K is a recently released large-scale tracking dataset containing 700 challenging test sequences, known for its diverse and difficult scenarios. On this demanding benchmark, our tracker again achieves state-of-the-art performance, as shown in Table~\ref{performance_architecture}. Notably, our \mytracker{}-B384 obtains an AUC of 63.0\%, surpassing strong competitors like SPMTrack-B384 (62.0\% AUC) by a significant margin.

\noindent\textbf{OTB100 \cite{otb100} and NfS \cite{nfs}.} OTB100 is a popular short-term tracking benchmark with 100 sequences covering 11 challenges like deformation and occlusion, while NfS contains 100 challenging, high-frame-rate videos. As shown in Table~\ref{tab:nfs_otb100}, our \mytracker{}-B224 surpasses all state-of-the-art trackers with significantly larger model sizes, demonstrating the superior efficiency and effectiveness of our design.

\subsection{Ablation Study and Analysis} 
In the ablation study, we will investigate the impact of our ATC module, HIBlock, number of template frames, different backbones, and varying token keep ratios ($r$) on model performance. All the ablation study is based on \mytracker{}-B224. Unless otherwise specified, we will default to $r$ value of 0.9 and 5 template frames.



\begin{table}[t!]
   \centering
   \setlength\tabcolsep{14.5pt} 
   \renewcommand{\arraystretch}{0.7} 
   \fontsize{8}{10.8}\selectfont
   \begin{tabular}{c|c c c}
   \toprule
   Number of Templates & AUC & P$_{norm}$ & P \\
   \midrule
   3 frames & 73.9 & 83.9 & 81.5 \\
   4 frames & 74.5 & 84.4 & 81.8 \\
   5 frames & 74.9 & 85.1 & 82.7 \\
   6 frames & 75.2 & 85.3 & 83.0 \\
   \bottomrule
   \end{tabular}
   \caption{Ablation on the Number of Templates on LaSOT.}
   \label{ablation_template_number}
\end{table}

\noindent\textbf{The Importance of ATC Module and HIBlock.}
Table~\ref{ablation_compression_interation} validates the significance of the two core components in our work: the ATC module and the HIBlock. In this study, removing the ATC module means all template tokens are used without redundancy elimination, while removing the HIBlock means replacing it with the standard backbone blocks \cite{itpn} for interaction. Our baseline model, where both ATC module and HIBlock are removed, processes all 5 template frames and the search region directly through the backbone before prediction.
The results on LaSOT are analyzed as follows (TNL2K results in Table~\ref{ablation_compression_interation}). Introducing only the ATC module to the baseline substantially improves performance by +0.7 AUC, demonstrating its high effectiveness in visual redundancy elimination. Adding only the HIBlock boosts the AUC score by +0.7, showcasing its powerful contextual modeling. When both modules are integrated, the ATC module provides a further +0.5 AUC gain on top of the HIBlock-equipped baseline, confirming the combined effect of our two components.



\begin{table}[t!]
   \centering
   \setlength\tabcolsep{2pt} 
   \renewcommand{\arraystretch}{0.9} 
   \fontsize{8}{10.8}\selectfont
   \begin{tabular}{c|c|cc|cc}
   \toprule
   \multicolumn{1}{c|}{\multirow{2}{*}{Backbone}} & 
   \multicolumn{1}{c|}{\multirow{2}{*}{Our method}} & 
   \multicolumn{2}{c|}{LaSOT} & 
   \multicolumn{2}{c}{TNL2K} \\
   \cline{3-6}
   & & AUC & P$_{norm}$ & AUC & P$_{norm}$ \\
   \midrule
   \multicolumn{1}{c|}{\multirow{2}{*}{ViT-B}} & \xmark & 69.9 & 79.9 & 58.4 & 76.1 \\
   \multicolumn{1}{c|}{} & \cmark & 73.4(+3.5) & 83.8(+3.9) & 60.6(+2.2) & 78.1(+2.0) \\
   \midrule
   \multicolumn{1}{c|}{\multirow{2}{*}{Fast-iTPN-B}} & \xmark & 72.8 & 83.1 & 59.4 & 77.0 \\
   \multicolumn{1}{c|}{} & \cmark & \textbf{74.9}(+2.1) & \textbf{85.1}(+2.0) & \textbf{61.3}(+1.9) & \textbf{78.8}(+1.8) \\
   \bottomrule
   \end{tabular}
   \caption{Backbone impact on LaSOT and TNL2K performance.}
   \label{ablation_backbone}
\end{table}


\noindent\textbf{The Number of Templates.}
As shown in Table~\ref{ablation_template_number} and Figure.\ref{ratio_frame}, we investigate the impact of varying the number of template frames on LaSOT. While adding more frames initially boosts performance (+1.3\% AUC gain from 3 to 6 frames), we observe that baseline models without our ATC module suffer a significant performance decline beyond 5 frames. This suggests that naively adding frames introduces detrimental redundant information. In contrast, our model with ATC avoids this issue, effectively extracting useful cues from more template frames.

\noindent\textbf{Variants of the TCM Architectures.}
We explored several architectures for the Token Correlation Module (TCM): ViT-B (trained from scratch or fine-tuned from MAE pre-trained weights \cite{mae}), a 8-layer Transformer encoder trained from scratch, and the first 8 layers of Fast-ITPN-B fine-tuned from its pre-trained weights. As shown in Table~\ref{TCM_ScoreProjector_Architecture}, the first 8 layers of Fast-ITPN-B (without additional model parameters) surpass both ViT-B variants and an 8-layer Transformer encoder on LaSOT in terms of either tracking accuracy or inference speed. This result indicates that a larger model scale is not the decisive factor for this task; instead, a more streamlined and well-tailored architecture demonstrates superior competitiveness.


\noindent\textbf{Effect of backbone.} Specifically, integrating our method with either ViT-B or Fast-iTPN-B results in significant improvements on the LaSOT and TNL2K. Notably, the Fast-iTPN-B equipped with our method achieves the best performance, reaching 74.9 AUC on LaSOT, as well as 61.3 AUC on TNL2K, validating its effectiveness and adaptability.

\begin{figure}[t!]
   \centering
   \includegraphics[width=0.9\linewidth]{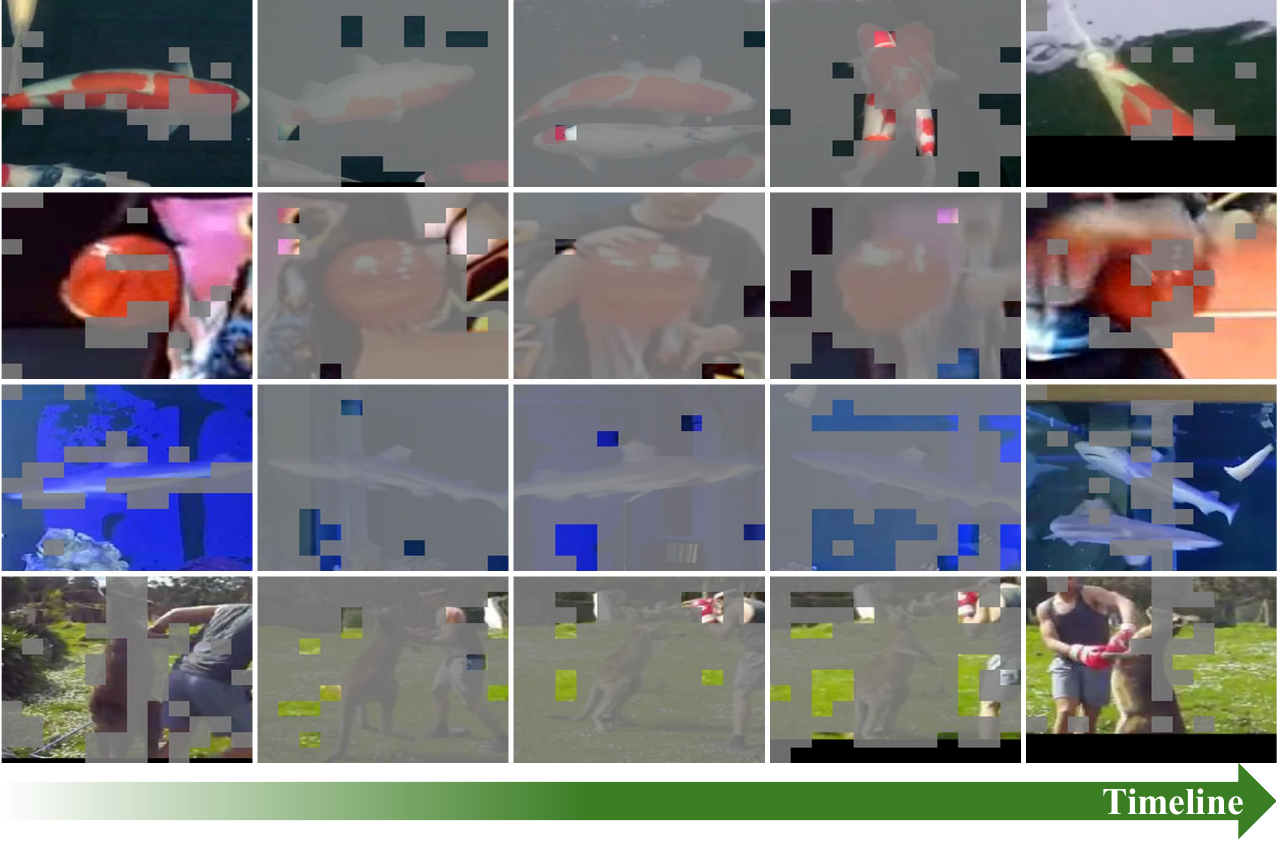}
   \caption{Visualization of visual redundancy elimination.}
   \label{redundant_tokens}
\end{figure}

\noindent\textbf{Visualization of token elimination.} Figure~ \ref{redundant_tokens} illustrates the token reduction process across sampled temporal templates. By employing the ATC module, \textbf{60\% of the total tokens} are effectively eliminated (represented by grey blocks) to prune redundant information while preserving informative features. Notably, the token elimination is more concentrated in the intermediate frames (2nd to 4th), which can be attributed to the high information redundancy following the TCM fusion; in these frames, a sparse subset of tokens is sufficient to represent the essential context. In contrast, the first and fifth frames retain more tokens, as the initial frame provides the most reliable target appearance and the final frame captures the latest dynamic cues, both of which are indispensable for robust tracking.

\noindent\textbf{Limitation.} 
While our approach achieves state-of-the-art performance with a very low computational cost, we believe there is a particularly exciting and impactful direction for future research. This valuable area of investigation is the development of fully dynamic token compression mechanisms, where the level of compression could adapt in real-time based on the complexity of the tracking scenario or the target's rapidly changing state.

\section{Conclusion}
In this work, we introduced \mytracker{}, a novel compress-then-interact visual tracking framework that effectively addresses visual redundancy in multi-frame Transformer trackers. Our framework consists of two key innovations: the \compressor{} module, which constructs a sparse, potent token set from historical template frames, and the Hierarchical Interaction Block, which performs a deep, multi-stage fusion of this compressed representation with the search region. Extensive experiments demonstrate that \mytracker{} achieves state-of-the-art performance on seven challenging benchmarks while maintaining a low computational cost. We believe these results validate that explicit information condensation is a promising direction for developing future high-performance trackers.

\section{Acknowledgements}
This work is supported by the Project of Guangxi Science and Technology (No. 2025GXNSFAA069676, 2025GXNSFAA069417, 2024GXNSFGA010001, and GuiKeFN2504240017), the National Natural Science Foundation of China (No.U23A20383, 62472109 and 62466051), the Guangxi ”Young Bagui Scholar” Teams for Innovation and Research Project, the Research Project of Guangxi Normal University (No. 2025DF001).

{
    \small
    \bibliographystyle{ieeenat_fullname}
    \bibliography{main}

@String(CVPR= {IEEE Conf. Comput. Vis. Pattern Recog.})

@String(ICCV= {Int. Conf. Comput. Vis.})

@String(ECCV= {Eur. Conf. Comput. Vis.})

@String(AAAI = {AAAI})

@String(CVPR  = {CVPR})

@String(ICCV  = {ICCV})

@String(ECCV  = {ECCV})

@inproceedings{SiamFC,
  title={Fully-convolutional siamese networks for object tracking},
  author={Bertinetto, Luca and Valmadre, Jack and Henriques, Joao F and Vedaldi, Andrea and Torr, Philip HS},
  booktitle={Computer vision--ECCV 2016 workshops: Amsterdam, the Netherlands, October 8-10 and 15-16, 2016, proceedings, part II 14},
  pages={850--865},
  year={2016},
  organization={Springer}
}

@inproceedings{SINT,
  title={Siamese instance search for tracking},
  author={Tao, Ran and Gavves, Efstratios and Smeulders, Arnold WM},
  booktitle={Proceedings of the IEEE conference on computer vision and pattern recognition},
  pages={1420--1429},
  year={2016}
}

@inproceedings{SiamRPN,
  title={High performance visual tracking with siamese region proposal network},
  author={Li, Bo and Yan, Junjie and Wu, Wei and Zhu, Zheng and Hu, Xiaolin},
  booktitle={Proceedings of the IEEE conference on computer vision and pattern recognition},
  pages={8971--8980},
  year={2018}
}

@article{attention_is_all_you_need,
  title={Attention is all you need},
  author={Vaswani, Ashish and Shazeer, Noam and Parmar, Niki and Uszkoreit, Jakob and Jones, Llion and Gomez, Aidan N and Kaiser, {\L}ukasz and Polosukhin, Illia},
  journal={Advances in neural information processing systems},
  volume={30},
  year={2017}
}

@inproceedings{TransT,
  title={Transformer tracking},
  author={Chen, Xin and Yan, Bin and Zhu, Jiawen and Wang, Dong and Yang, Xiaoyun and Lu, Huchuan},
  booktitle={Proceedings of the IEEE/CVF conference on computer vision and pattern recognition},
  pages={8126--8135},
  year={2021}
}

@inproceedings{Mixformer,
  title={Mixformer: End-to-end tracking with iterative mixed attention},
  author={Cui, Yutao and Jiang, Cheng and Wang, Limin and Wu, Gangshan},
  booktitle={Proceedings of the IEEE/CVF conference on computer vision and pattern recognition},
  pages={13608--13618},
  year={2022}
}

@inproceedings{OSTrack,
  title={Joint feature learning and relation modeling for tracking: A one-stream framework},
  author={Ye, Botao and Chang, Hong and Ma, Bingpeng and Shan, Shiguang and Chen, Xilin},
  booktitle={European conference on computer vision},
  pages={341--357},
  year={2022},
  organization={Springer}
}

@inproceedings{ODTrack,
  title={Odtrack: Online dense temporal token learning for visual tracking},
  author={Zheng, Yaozong and Zhong, Bineng and Liang, Qihua and Mo, Zhiyi and Zhang, Shengping and Li, Xianxian},
  booktitle={Proceedings of the AAAI conference on artificial intelligence},
  volume={38},
  number={7},
  pages={7588--7596},
  year={2024}
}

@inproceedings{TATrack,
  title={Target-aware tracking with long-term context attention},
  author={He, Kaijie and Zhang, Canlong and Xie, Sheng and Li, Zhixin and Wang, Zhiwen},
  booktitle={Proceedings of the AAAI conference on artificial intelligence},
  volume={37},
  number={1},
  pages={773--780},
  year={2023}
}

@inproceedings{VideoTrack,
  title={Videotrack: Learning to track objects via video transformer},
  author={Xie, Fei and Chu, Lei and Li, Jiahao and Lu, Yan and Ma, Chao},
  booktitle={Proceedings of the IEEE/CVF conference on computer vision and pattern recognition},
  pages={22826--22835},
  year={2023}
}

@article{SwinTrack,
  title={Swintrack: A simple and strong baseline for transformer tracking},
  author={Lin, Liting and Fan, Heng and Zhang, Zhipeng and Xu, Yong and Ling, Haibin},
  journal={Advances in Neural Information Processing Systems},
  volume={35},
  pages={16743--16754},
  year={2022}
}

@article{Video-XL-Pro,
  title={Video-xl-pro: Reconstructive token compression for extremely long video understanding},
  author={Liu, Xiangrui and Shu, Yan and Liu, Zheng and Li, Ao and Tian, Yang and Zhao, Bo},
  journal={arXiv preprint arXiv:2503.18478},
  year={2025}
}

@article{InternVL-X,
  title={InternVL-X: Advancing and Accelerating InternVL Series with Efficient Visual Token Compression},
  author={Lu, Dongchen and Sun, Yuyao and Zhang, Zilu and Huang, Leping and Zeng, Jianliang and Shu, Mao and Cao, Huo},
  journal={arXiv preprint arXiv:2503.21307},
  year={2025}
}

@article{FocusLLaVA,
  title={FocusLLaVA: A Coarse-to-Fine Approach for Efficient and Effective Visual Token Compression},
  author={Zhu, Yuke and Xie, Chi and Liang, Shuang and Zheng, Bo and Guo, Sheng},
  journal={arXiv preprint arXiv:2411.14228},
  year={2024}
}

@inproceedings{Hybrid-Level-Instruction-Injection,
  title={Hybrid-level instruction injection for video token compression in multi-modal large language models},
  author={Liu, Zhihang and Xie, Chen-Wei and Li, Pandeng and Zhao, Liming and Tang, Longxiang and Zheng, Yun and Liu, Chuanbin and Xie, Hongtao},
  booktitle={Proceedings of the Computer Vision and Pattern Recognition Conference},
  pages={8568--8578},
  year={2025}
}

@inproceedings{fan2019siamese,
  title={Siamese cascaded region proposal networks for real-time visual tracking},
  author={Fan, Heng and Ling, Haibin},
  booktitle={Proceedings of the IEEE/CVF conference on computer vision and pattern recognition},
  pages={7952--7961},
  year={2019}
}

@inproceedings{li2019siamrpn++,
  title={Siamrpn++: Evolution of siamese visual tracking with very deep networks},
  author={Li, Bo and Wu, Wei and Wang, Qiang and Zhang, Fangyi and Xing, Junliang and Yan, Junjie},
  booktitle={Proceedings of the IEEE/CVF conference on computer vision and pattern recognition},
  pages={4282--4291},
  year={2019}
}

@inproceedings{li2018high,
  title={High performance visual tracking with siamese region proposal network},
  author={Li, Bo and Yan, Junjie and Wu, Wei and Zhu, Zheng and Hu, Xiaolin},
  booktitle={Proceedings of the IEEE conference on computer vision and pattern recognition},
  pages={8971--8980},
  year={2018}
}

@inproceedings{fu2021stmtrack,
  title={Stmtrack: Template-free visual tracking with space-time memory networks},
  author={Fu, Zhihong and Liu, Qingjie and Fu, Zehua and Wang, Yunhong},
  booktitle={Proceedings of the IEEE/CVF conference on computer vision and pattern recognition},
  pages={13774--13783},
  year={2021}
}

@article{vit,
  title={An image is worth 16x16 words: Transformers for image recognition at scale},
  author={Dosovitskiy, Alexey and Beyer, Lucas and Kolesnikov, Alexander and Weissenborn, Dirk and Zhai, Xiaohua and Unterthiner, Thomas and Dehghani, Mostafa and Minderer, Matthias and Heigold, Georg and Gelly, Sylvain and others},
  journal={arXiv preprint arXiv:2010.11929},
  year={2020}
}

@inproceedings{hivit,
  title={Hivit: A simpler and more efficient design of hierarchical vision transformer},
  author={Zhang, Xiaosong and Tian, Yunjie and Xie, Lingxi and Huang, Wei and Dai, Qi and Ye, Qixiang and Tian, Qi},
  booktitle={The Eleventh International Conference on Learning Representations},
  year={2023}
}

@article{itpn,
  title={Fast-iTPN: Integrally pre-trained transformer pyramid network with token migration},
  author={Tian, Yunjie and Xie, Lingxi and Qiu, Jihao and Jiao, Jianbin and Wang, Yaowei and Tian, Qi and Ye, Qixiang},
  journal={IEEE Transactions on Pattern Analysis and Machine Intelligence},
  year={2024},
  publisher={IEEE}
}

@inproceedings{swintransformer,
  title={Swin transformer: Hierarchical vision transformer using shifted windows},
  author={Liu, Ze and Lin, Yutong and Cao, Yue and Hu, Han and Wei, Yixuan and Zhang, Zheng and Lin, Stephen and Guo, Baining},
  booktitle={Proceedings of the IEEE/CVF international conference on computer vision},
  pages={10012--10022},
  year={2021}
}

@inproceedings{artrack,
  title={Autoregressive visual tracking},
  author={Wei, Xing and Bai, Yifan and Zheng, Yongchao and Shi, Dahu and Gong, Yihong},
  booktitle={Proceedings of the IEEE/CVF Conference on Computer Vision and Pattern Recognition},
  pages={9697--9706},
  year={2023}
}

@inproceedings{evptrack,
  title={Explicit visual prompts for visual object tracking},
  author={Shi, Liangtao and Zhong, Bineng and Liang, Qihua and Li, Ning and Zhang, Shengping and Li, Xianxian},
  booktitle={Proceedings of the AAAI Conference on Artificial Intelligence},
  volume={38},
  number={5},
  pages={4838--4846},
  year={2024}
}

@inproceedings{artrackv2,
  title={Artrackv2: Prompting autoregressive tracker where to look and how to describe},
  author={Bai, Yifan and Zhao, Zeyang and Gong, Yihong and Wei, Xing},
  booktitle={Proceedings of the IEEE/CVF conference on computer vision and pattern recognition},
  pages={19048--19057},
  year={2024}
}

@inproceedings{blip2,
  title={Blip-2: Bootstrapping language-image pre-training with frozen image encoders and large language models},
  author={Li, Junnan and Li, Dongxu and Savarese, Silvio and Hoi, Steven},
  booktitle={International conference on machine learning},
  pages={19730--19742},
  year={2023},
  organization={PMLR}
}

@article{flamingo,
  title={Flamingo: a visual language model for few-shot learning},
  author={Alayrac, Jean-Baptiste and Donahue, Jeff and Luc, Pauline and Miech, Antoine and Barr, Iain and Hasson, Yana and Lenc, Karel and Mensch, Arthur and Millican, Katherine and Reynolds, Malcolm and others},
  journal={Advances in neural information processing systems},
  volume={35},
  pages={23716--23736},
  year={2022}
}

@article{matryoshka,
  title={Matryoshka query transformer for large vision-language models},
  author={Hu, Wenbo and Dou, Zi-Yi and Li, Liunian and Kamath, Amita and Peng, Nanyun and Chang, Kai-Wei},
  journal={Advances in Neural Information Processing Systems},
  volume={37},
  pages={50168--50188},
  year={2024}
}

@article{Moe-llava,
  title={Moe-llava: Mixture of experts for large vision-language models},
  author={Lin, Bin and Tang, Zhenyu and Ye, Yang and Cui, Jiaxi and Zhu, Bin and Jin, Peng and Huang, Jinfa and Zhang, Junwu and Pang, Yatian and Ning, Munan and others},
  journal={arXiv preprint arXiv:2401.15947},
  year={2024}
}

@inproceedings{liu2024improved,
  title={Improved baselines with visual instruction tuning},
  author={Liu, Haotian and Li, Chunyuan and Li, Yuheng and Lee, Yong Jae},
  booktitle={Proceedings of the IEEE/CVF Conference on Computer Vision and Pattern Recognition},
  pages={26296--26306},
  year={2024}
}

@inproceedings{Madtp,
  title={Madtp: Multimodal alignment-guided dynamic token pruning for accelerating vision-language transformer},
  author={Cao, Jianjian and Ye, Peng and Li, Shengze and Yu, Chong and Tang, Yansong and Lu, Jiwen and Chen, Tao},
  booktitle={Proceedings of the IEEE/CVF conference on computer vision and pattern recognition},
  pages={15710--15719},
  year={2024}
}

@article{Otterhd,
  title={Otterhd: A high-resolution multi-modality model},
  author={Li, Bo and Zhang, Peiyuan and Yang, Jingkang and Zhang, Yuanhan and Pu, Fanyi and Liu, Ziwei},
  journal={arXiv preprint arXiv:2311.04219},
  year={2023}
}

@article{Llava-prumerge,
  title={Llava-prumerge: Adaptive token reduction for efficient large multimodal models},
  author={Shang, Yuzhang and Cai, Mu and Xu, Bingxin and Lee, Yong Jae and Yan, Yan},
  journal={arXiv preprint arXiv:2403.15388},
  year={2024}
}

@article{zhang2024tokenlevel,
  title={Token-level correlation-guided compression for efficient multimodal document understanding},
  author={Zhang, Renshan and Lyu, Yibo and Shao, Rui and Chen, Gongwei and Guan, Weili and Nie, Liqiang},
  journal={arXiv preprint arXiv:2407.14439},
  year={2024}
}

@inproceedings{LMTrack,
  title={Less is more: Token context-aware learning for object tracking},
  author={Xu, Chenlong and Zhong, Bineng and Liang, Qihua and Zheng, Yaozong and Li, Guorong and Song, Shuxiang},
  booktitle={Proceedings of the AAAI Conference on Artificial Intelligence},
  volume={39},
  number={8},
  pages={8824--8832},
  year={2025}
}

@inproceedings{focal_loss,  
title={Focal Loss for Dense Object Detection}, 
url={http://dx.doi.org/10.1109/iccv.2017.324}, 
DOI={10.1109/iccv.2017.324}, 
booktitle={2017 IEEE International Conference on Computer Vision (ICCV)}, 
author={Lin, Tsung-Yi and Goyal, Priya and Girshick, Ross and He, Kaiming and Dollar, Piotr}, 
year={2017}, 
month={Oct}, 
language={en-US} 
}

@inproceedings{giou,  
title={Generalized Intersection over Union: A Metric and A Loss for Bounding Box Regression}, 
url={http://dx.doi.org/10.1109/cvpr.2019.00075}, 
DOI={10.1109/cvpr.2019.00075}, 
booktitle={2019 IEEE/CVF Conference on Computer Vision and Pattern Recognition (CVPR)}, 
author={Rezatofighi, Hamid and Tsoi, Nathan and Gwak, JunYoung and Sadeghian, Amir and Reid, Ian and Savarese, Silvio}, 
year={2019}, 
month={Jun}, 
language={en-US} 
}

@article{got10k,  
title={GOT-10k: A Large High-Diversity Benchmark for Generic Object Tracking in the Wild}, 
url={http://dx.doi.org/10.1109/tpami.2019.2957464}, 
DOI={10.1109/tpami.2019.2957464}, 
journal={IEEE Transactions on Pattern Analysis and Machine Intelligence}, 
author={Huang, Lianghua and Zhao, Xin and Huang, Kaiqi}, 
year={2021}, 
month={May}, 
pages={1562–1577}, 
language={en-US} 
}

@inproceedings{lasot,  
title={LaSOT: A High-quality Benchmark for Large-scale Single Object Tracking}, 
url={http://dx.doi.org/10.1109/cvpr.2019.00552}, 
DOI={10.1109/cvpr.2019.00552}, 
booktitle={2019 IEEE/CVF Conference on Computer Vision and Pattern Recognition (CVPR)}, 
author={Fan, Heng and Lin, Liting and Yang, Fan and Chu, Peng and Deng, Ge and Yu, Sijia and Bai, Hexin and Xu, Yong and Liao, Chunyuan and Ling, Haibin}, 
year={2019}, 
month={Jun}, 
language={en-US} 
}

@inbook{trackingnet,  
title={TrackingNet: A Large-Scale Dataset and Benchmark for Object Tracking in the Wild}, 
url={http://dx.doi.org/10.1007/978-3-030-01246-5_19}, 
DOI={10.1007/978-3-030-01246-5_19}, 
booktitle={Computer Vision – ECCV 2018,Lecture Notes in Computer Science}, 
author={Müller, Matthias and Bibi, Adel and Giancola, Silvio and Alsubaihi, Salman and Ghanem, Bernard}, 
year={2018}, 
month={Jan}, 
pages={310–327}, 
language={en-US} 
}

@inbook{coco,  
title={Microsoft COCO: Common Objects in Context}, 
url={http://dx.doi.org/10.1007/978-3-319-10602-1_48}, 
DOI={10.1007/978-3-319-10602-1_48}, 
booktitle={Computer Vision – ECCV 2014,Lecture Notes in Computer Science}, 
author={Lin, Tsung-Yi and Maire, Michael and Belongie, Serge and Hays, James and Perona, Pietro and Ramanan, Deva and Dollár, Piotr and Zitnick, C. Lawrence}, 
year={2014}, 
month={Jan}, 
pages={740–755}, 
language={en-US} 
}

@article{vastrack,
  title={Vasttrack: Vast category visual object tracking},
  author={Peng, Liang and Gao, Junyuan and Liu, Xinran and Li, Weihong and Dong, Shaohua and Zhang, Zhipeng and Fan, Heng and Zhang, Libo},
  journal={Advances in Neural Information Processing Systems},
  volume={37},
  pages={130797--130818},
  year={2024}
}

@article{lasot_ext,  
title={LaSOT: A High-quality Large-scale Single Object Tracking Benchmark}, 
journal={International Journal of Computer Vision,International Journal of Computer Vision}, 
author={Fan, Heng and Bai, Hexin and Lin, Liting and Yang, Fan and Chu, Peng and Deng, Ge and Yu, Sijia and Harshit, Harshit and Huang, Mingzhen and Liu, Juehuan and Xu, Yong and Liao, Chunyuan and Lin, Yuan and Ling, Haibin}, 
year={2020}, 
month={Sep}, 
language={en-US} 
}

@inproceedings{tnl2k,
  title={Towards more flexible and accurate object tracking with natural language: Algorithms and benchmark},
  author={Wang, Xiao and Shu, Xiujun and Zhang, Zhipeng and Jiang, Bo and Wang, Yaowei and Tian, Yonghong and Wu, Feng},
  booktitle={Proceedings of the IEEE/CVF conference on computer vision and pattern recognition},
  pages={13763--13773},
  year={2021}
}

@inproceedings{nfs,
  title={Need for speed: A benchmark for higher frame rate object tracking},
  author={Kiani Galoogahi, Hamed and Fagg, Ashton and Huang, Chen and Ramanan, Deva and Lucey, Simon},
  booktitle={Proceedings of the IEEE international conference on computer vision},
  pages={1125--1134},
  year={2017}
}

@article{otb100,  
title={Object Tracking Benchmark}, 
url={http://dx.doi.org/10.1109/tpami.2014.2388226}, 
DOI={10.1109/tpami.2014.2388226}, 
journal={IEEE Transactions on Pattern Analysis and Machine Intelligence}, 
author={Wu, Yi and Lim, Jongwoo and Yang, Ming-Hsuan}, 
year={2015}, 
month={Sep}, 
pages={1834–1848}, 
language={en-US} 
}

@inproceedings{F-BDMTrack,
  title={Foreground-background distribution modeling transformer for visual object tracking},
  author={Yang, Dawei and He, Jianfeng and Ma, Yinchao and Yu, Qianjin and Zhang, Tianzhu},
  booktitle={Proceedings of the IEEE/CVF international conference on computer vision},
  pages={10117--10127},
  year={2023}
}

@article{seqtrack,  
title={SeqTrack: Sequence to Sequence Learning for Visual Object Tracking}, 
author={Chen, Xin and Peng, Houwen and Wang, Dong and Lu, Huchuan and Hu, Han}, 
language={en-US} 
}

@inproceedings{STark,  
title={Learning Spatio-Temporal Transformer for Visual Tracking}, 
url={http://dx.doi.org/10.1109/iccv48922.2021.01028}, 
DOI={10.1109/iccv48922.2021.01028}, 
booktitle={2021 IEEE/CVF International Conference on Computer Vision (ICCV)}, 
author={Yan, Bin and Peng, Houwen and Fu, Jianlong and Wang, Dong and Lu, Huchuan}, 
year={2021}, 
month={Oct}, 
language={en-US} 
}

@InProceedings{hiptrack,
   author    = {Cai, Wenrui and Liu, Qingjie and Wang, Yunhong},
   title     = {HIPTrack: Visual Tracking with Historical Prompts},
   booktitle = {Proceedings of the IEEE/CVF Conference on Computer Vision and Pattern Recognition (CVPR)},
   month     = {June},
   year      = {2024},
   pages     = {19258-19267}
}

@inproceedings{VideoTrack1,
  title={Videotrack: Learning to track objects via video transformer},
  author={Xie, Fei and Chu, Lei and Li, Jiahao and Lu, Yan and Ma, Chao},
  booktitle={Proceedings of the IEEE/CVF conference on computer vision and pattern recognition},
  pages={22826--22835},
  year={2023}
}

@inproceedings{temtrack,
  title={Robust tracking via mamba-based context-aware token learning},
  author={Xie, Jinxia and Zhong, Bineng and Liang, Qihua and Li, Ning and Mo, Zhiyi and Song, Shuxiang},
  booktitle={Proceedings of the AAAI Conference on Artificial Intelligence},
  volume={39},
  number={8},
  pages={8727--8735},
  year={2025}
}

@inproceedings{spmtrack,
  title={SPMTrack: Spatio-Temporal Parameter-Efficient Fine-Tuning with Mixture of Experts for Scalable Visual Tracking},
  author={Cai, Wenrui and Liu, Qingjie and Wang, Yunhong},
  booktitle={Proceedings of the Computer Vision and Pattern Recognition Conference},
  pages={16871--16881},
  year={2025}
}

@inproceedings{ARPTrack,
  title={Autoregressive Sequential Pretraining for Visual Tracking},
  author={Liang, Shiyi and Bai, Yifan and Gong, Yihong and Wei, Xing},
  booktitle={Proceedings of the Computer Vision and Pattern Recognition Conference},
  pages={7254--7264},
  year={2025}
}

@inproceedings{DreamTrack,
  title={DreamTrack: Dreaming the Future for Multimodal Visual Object Tracking},
  author={Guo, Mingzhe and Tan, Weiping and Ran, Wenyu and Jing, Liping and Zhang, Zhipeng},
  booktitle={Proceedings of the Computer Vision and Pattern Recognition Conference},
  pages={7201--7210},
  year={2025}
}

@inproceedings{MCITrack,
  title={Exploring enhanced contextual information for video-level object tracking},
  author={Kang, Ben and Chen, Xin and Lai, Simiao and Liu, Yang and Liu, Yi and Wang, Dong},
  booktitle={Proceedings of the AAAI Conference on Artificial Intelligence},
  volume={39},
  number={4},
  pages={4194--4202},
  year={2025}
}

@inproceedings{mae,
  title={Masked autoencoders are scalable vision learners},
  author={He, Kaiming and Chen, Xinlei and Xie, Saining and Li, Yanghao and Doll{\'a}r, Piotr and Girshick, Ross},
  booktitle={Proceedings of the IEEE/CVF conference on computer vision and pattern recognition},
  pages={16000--16009},
  year={2022}
}

@article{nvidia2025_token_efficient,
  title={Token-efficient long video understanding for multimodal llms},
  author={Jiang, Jindong and Li, Xiuyu and Liu, Zhijian and Li, Muyang and Chen, Guo and Li, Zhiqi and Huang, De-An and Liu, Guilin and Yu, Zhiding and Keutzer, Kurt and others},
  journal={arXiv preprint arXiv:2503.04130},
  year={2025}
}

@inproceedings{hu2025exploiting,
  title={Exploiting multimodal spatial-temporal patterns for video object tracking},
  author={Hu, Xiantao and Tai, Ying and Zhao, Xu and Zhao, Chen and Zhang, Zhenyu and Li, Jun and Zhong, Bineng and Yang, Jian},
  booktitle={Proceedings of the AAAI Conference on Artificial Intelligence},
  volume={39},
  number={4},
  pages={3581--3589},
  year={2025}
}

@article{li2025cadtrack,
  title={CADTrack: Learning Contextual Aggregation with Deformable Alignment for Robust RGBT Tracking},
  author={Li, Hao and Wang, Yuhao and Hu, Xiantao and Hao, Wenning and Zhang, Pingping and Wang, Dong and Lu, Huchuan},
  journal={arXiv preprint arXiv:2511.17967},
  year={2025}
}

@article{hu2025adaptive,
  title={Adaptive perception for unified visual multi-modal object tracking},
  author={Hu, Xiantao and Zhong, Bineng and Liang, Qihua and Shi, Liangtao and Mo, Zhiyi and Tai, Ying and Yang, Jian},
  journal={IEEE Transactions on Artificial Intelligence},
  year={2025},
  publisher={IEEE}
}

@article{hu2023transformer,
  title={Transformer tracking via frequency fusion},
  author={Hu, Xiantao and Zhong, Bineng and Liang, Qihua and Zhang, Shengping and Li, Ning and Li, Xianxian and Ji, Rongrong},
  journal={IEEE Transactions on Circuits and Systems for Video Technology},
  volume={34},
  number={2},
  pages={1020--1031},
  year={2023},
  publisher={IEEE}
}

@article{hu2024toward,
  title={Toward modalities correlation for RGB-T tracking},
  author={Hu, Xiantao and Zhong, Bineng and Liang, Qihua and Zhang, Shengping and Li, Ning and Li, Xianxian},
  journal={IEEE Transactions on Circuits and Systems for Video Technology},
  volume={34},
  number={10},
  pages={9102--9111},
  year={2024},
  publisher={IEEE}
}

@inproceedings{xue2025similarity,
  title={Similarity-guided layer-adaptive vision transformer for UAV tracking},
  author={Xue, Chaocan and Zhong, Bineng and Liang, Qihua and Zheng, Yaozong and Li, Ning and Xue, Yuanliang and Song, Shuxiang},
  booktitle={Proceedings of the Computer Vision and Pattern Recognition Conference},
  pages={6730--6740},
  year={2025}
}

@inproceedings{song1,
  title={Transformer tracking with cyclic shifting window attention},
  author={Song, Zikai and Yu, Junqing and Chen, Yi-Ping Phoebe and Yang, Wei},
  booktitle={Proceedings of the IEEE/CVF conference on computer vision and pattern recognition},
  pages={8791--8800},
  year={2022}
}

@inproceedings{song2,
  title={Compact transformer tracker with correlative masked modeling},
  author={Song, Zikai and Luo, Run and Yu, Junqing and Chen, Yi-Ping Phoebe and Yang, Wei},
  booktitle={Proceedings of the AAAI conference on artificial intelligence},
  volume={37},
  number={2},
  pages={2321--2329},
  year={2023}
}

@article{zheng2025towards,
  title={Towards universal modal tracking with online dense temporal token learning},
  author={Zheng, Yaozong and Zhong, Bineng and Liang, Qihua and Zhang, Shengping and Li, Guorong and Li, Xianxian and Ji, Rongrong},
  journal={IEEE Transactions on Pattern Analysis and Machine Intelligence},
  year={2025},
  publisher={IEEE}
}

@article{gao2022robust,
  title={Robust tracking via learning model update with unsupervised anomaly detection philosophy},
  author={Gao, Jie and Zhong, Bineng and Chen, Yan},
  journal={IEEE Transactions on Circuits and Systems for Video Technology},
  volume={33},
  number={5},
  pages={2330--2341},
  year={2022},
  publisher={IEEE}
}

@inproceedings{gao2023unambiguous,
  title={Unambiguous object tracking by exploiting target cues},
  author={Gao, Jie and Zhong, Bineng and Chen, Yan},
  booktitle={Proceedings of the 31st ACM international conference on multimedia},
  pages={1997--2005},
  year={2023}
}

@inproceedings{sstrack,
  title={Decoupled Spatio-Temporal Consistency Learning for Self-Supervised Tracking},
  author={Zheng, Yaozong and Zhong, Bineng and Liang, Qihua and Li, Ning and Song, Shuxiang},
  booktitle={Proceedings of the AAAI Conference on Artificial Intelligence},
  volume={39},
  number={10},
  pages={10635--10643},
  year={2025}
}

@article{MMTrack,
  title={Toward unified token learning for vision-language tracking},
  author={Zheng, Yaozong and Zhong, Bineng and Liang, Qihua and Li, Guorong and Ji, Rongrong and Li, Xianxian},
  journal={IEEE Transactions on Circuits and Systems for Video Technology},
  volume={34},
  number={4},
  pages={2125--2135},
  year={2023},
  publisher={IEEE}
}

@article{SiamPIN,
  title={Leveraging local and global cues for visual tracking via parallel interaction network},
  author={Zheng, Yaozong and Zhong, Bineng and Liang, Qihua and Tang, Zhenjun and Ji, Rongrong and Li, Xianxian},
  journal={IEEE Transactions on Circuits and Systems for Video Technology},
  volume={33},
  number={4},
  pages={1671--1683},
  year={2022},
  publisher={IEEE}
}
}


\end{document}